\documentclass[letterpaper]{article} % DO NOT CHANGE THIS
\usepackage{aaai2026}  % DO NOT CHANGE THIS
\usepackage{times}  % DO NOT CHANGE THIS
\usepackage{helvet}  % DO NOT CHANGE THIS
\usepackage{courier}  % DO NOT CHANGE THIS
\usepackage[hyphens]{url}  % DO NOT CHANGE THIS
\usepackage{graphicx} % DO NOT CHANGE THIS
\usepackage{natbib}  % DO NOT CHANGE THIS AND DO NOT ADD ANY OPTIONS TO IT
\usepackage{caption} % DO NOT CHANGE THIS AND DO NOT ADD ANY OPTIONS TO IT
\usepackage{algorithm}
\usepackage{algorithmic}

\usepackage{amsmath,amsfonts}
\usepackage{bm}
\usepackage{ragged2e}
\usepackage{color}
\usepackage{multirow}
\usepackage{threeparttable}
\usepackage{booktabs}
\usepackage{cuted}
\usepackage{tikz}

\usepackage{subfigure}
\usepackage{graphicx}
\usepackage{array}
\usepackage{colortbl}
\usepackage{newfloat}
\usepackage{listings}
\DeclareCaptionStyle{ruled}{labelfont=normalfont,labelsep=colon,strut=off} % DO NOT CHANGE THIS
\floatstyle{ruled}
\newfloat{listing}{tb}{lst}{}
\floatname{listing}{Listing}
\title{QueryCraft: Transformer-Guided Query Initialization for Enhanced Human-Object Interaction Detection}
\author{
    Yuxiao Wang\textsuperscript{\rm 1}\equalcontrib, Wolin Liang\textsuperscript{\rm 1}\equalcontrib, Yu Lei\textsuperscript{\rm 2}, Weiying Xue\textsuperscript{\rm 1}, Nan Zhuang\textsuperscript{\rm 3}, Qi Liu\textsuperscript{\rm 1}\thanks{Corresponding author}
}
\affiliations{
    \textsuperscript{\rm 1}School of Future Technology, South China University of Technology\\
    \textsuperscript{\rm 2}School of Information Science \& Technology, Southwest Jiaotong University\\
    \textsuperscript{\rm 3}School of Software Technology, Zhejiang University

}

\usepackage{bibentry}
\begin{document}

\maketitle

\begin{abstract}
% 人-物交互（Human-Object Interaction, HOI）检测旨在定位图像中的人-物体对并识别其交互关系。近年来，虽然基于 DETR 的方法逐渐成为主流，但依然存在一些可以改进的问题：其一，缺乏外部语义知识指导，导致交互语义理解能力有限；其二，查询初始化语义不明确，影响整体检测性能。

% 为解决上述问题，本文提出一种融合语义对齐与先验约束的 HOI 检测方法。首先，我们引入双 CLIP 编码结构：CLIP-Integrated Interaction Transformer（CLINT） ，将图像与文本描述对齐至统一语义空间，通过 Transformer 编码器深入建模交互语义，提升交互理解能力。其次，为缓解查询语义模糊问题，我们设计了一个目标感知模块：Perceptual Distilled Query Decoder (PDQD)，该模块不仅负责物体语义特征的学习，还引入了 YOLO 模型提供的目标检测结果，在训练阶段构造辅助监督损失，以增强 decoder 对图像中物体类别的感知能力。最终，从 decoder 中提取的语义特征用于初始化物体查询（object queries），实现语义明确的 query 表达。

% 我们的方法在 HICO-Det 和 V-COCO 数据集上取得了优越的检测性能，验证了双重语义增强与先验引导机制在 HOI 任务中的有效性与通用性。

Human-Object Interaction (HOI) detection aims to localize human-object pairs and recognize their interactions in images. Although DETR-based methods have recently emerged as the mainstream framework for HOI detection, they still suffer from a key limitation: Randomly initialized queries lack explicit semantics, leading to suboptimal detection performance. To address this challenge, we propose QueryCraft, a novel plug-and-play HOI detection framework that incorporates semantic priors and guided feature learning through transformer-based query initialization. Central to our approach is \textbf{ACTOR} (\textbf{A}ction-aware \textbf{C}ross-modal \textbf{T}ransf\textbf{OR}mer), a cross-modal Transformer encoder that jointly attends to visual regions and textual prompts to extract action-relevant features. Rather than merely aligning modalities, ACTOR leverages language-guided attention to infer interaction semantics and produce semantically meaningful query representations. To further enhance object-level query quality, we introduce a \textbf{P}erceptual \textbf{D}istilled \textbf{Q}uery \textbf{D}ecoder (\textbf{PDQD}), which distills object category awareness from a pre-trained detector to serve as object query initiation. This dual-branch query initialization enables the model to generate more interpretable and effective queries for HOI detection. Extensive experiments on HICO-Det and V-COCO benchmarks demonstrate that our method achieves state-of-the-art performance and strong generalization. Code will be released upon publication.

\end{abstract}


\section{introduction}

\begin{figure}[ht]
\centering %表示居中
\includegraphics[width=\linewidth]{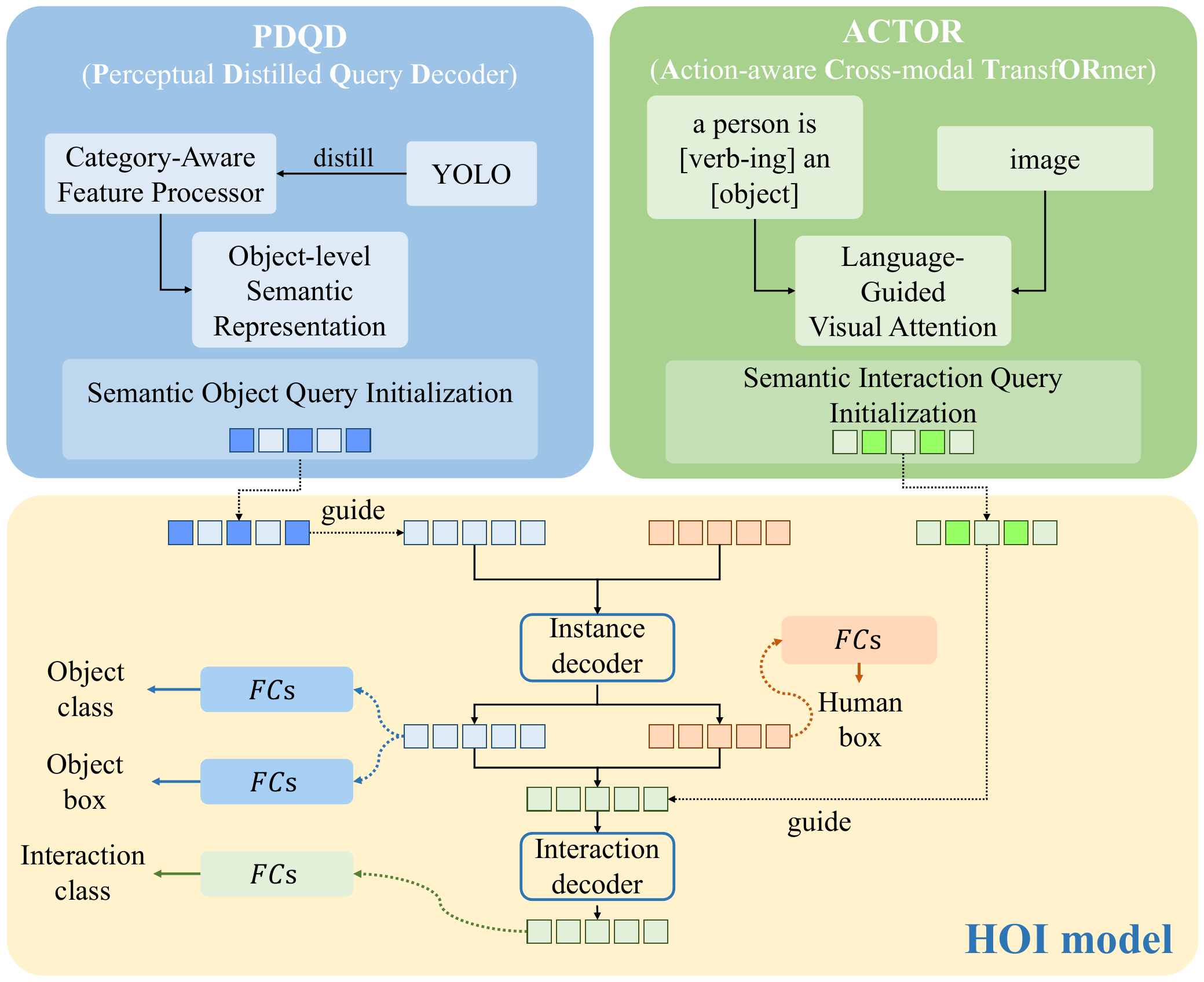}
% [height=4.5cm]表示高度
%[width=9.5cm]表示宽度
%{111.eps}表示eps格式的图片，名为111

\caption{Overview of QueryCraft. PDQD (left) produces object-level semantic representations by distilling knowledge from YOLO. ACTOR (right) generates semantic interaction queries through language-guided visual attention using text prompts. These semantically-enriched queries guide the Instance and Interaction Decoders to predict human boxes, object boxes/classes, and interaction classes, forming the complete HOI detection model.
% In addition to the HOI model, our model comprises three vital modules. The Candidate Image Construction module extracts humans and objects, performs segmentation, and establishes one-to-one matching of human-object. 
}
%图片的名称
\label{fig:Figure_1}
%图片的标签，用于文章中的引用，注意到标签的数字与实际文章显示的数字可能不同
% \vspace{-0.2cm}
\end{figure}

Human-Object Interaction (HOI) detection aims to identify and localize human-object pairs in images while recognizing their interactions, generating structured $\langle$human, action, object$\rangle$ triplets. Beyond traditional detection tasks, HOI detection requires understanding semantic relationships between entities, demanding robust scene modeling capabilities. With applications spanning behavior recognition, image captioning, video analysis, and robotic perception, this task has attracted significant research attention~\cite{liao2022gen,wang2025pihot}.
% Human-Object Interaction (HOI) detection is a core task in computer vision, which aims to localize humans and the objects they interact with in an image, and to recognize the interaction between them, thereby generating structured triplets in the form of ⟨human, action, object⟩. Unlike traditional object detection and image classification, HOI detection not only requires accurate recognition of individual entities in the image but also demands understanding of the semantic relationships between them, necessitating strong scene modeling and reasoning capabilities. This task has wide applications in behavior recognition, image captioning, video analysis, human-computer interaction, and robotic perception, and has thus garnered increasing attention from the research community in recent years.

Early HOI detection methods follow a two-stage pipeline: detecting humans and objects via detectors like Faster R-CNN~\cite{ren2015faster}, then classifying interactions through feature fusion, spatial encodings, or graph-based reasoning~\cite{chao2018learning,wang2024dehot,tip-9552553}. Despite reasonable performance, these approaches suffer from computational inefficiency due to combinatorial pair explosion and limited end-to-end optimization capability~\cite{wang2024review}. The emergence of Transformer-based architectures, particularly DETR's set prediction formulation, has revolutionized HOI detection by eliminating redundant proposals. DETR-inspired methods like GEN-VLKT~\cite{liao2022gen}, TED-Net~\cite{wang2024ted}, and KI2HOI~\cite{xue2025towards} have advanced the field through interaction-aware queries, action embeddings, and spatial priors, significantly improving detection performance. Despite advances in DETR-based HOI detection, key challenges remain in modeling complex interactions and semantic generalization~\cite{yuan2023rlipv2,wang2024freea}. DETR's random query initialization lacks explicit semantic grounding, limiting accurate representation of object categories and interaction semantics. 

To address the limitations of randomly initialized queries in DETR-based HOI detection, we propose QueryCraft, a novel framework that incorporates semantic priors and guided feature learning through transformer-based query initialization, as shown in Figure~\ref{fig:Figure_1}. At the core of our approach is ACTOR (Action-aware Cross-modal TransfORmer), a cross-modal transformer encoder that jointly leverages semantic alignment and prior-guided mechanisms to enhance both the semantic understanding of interactions and the clarity of query representations. 
% ACTOR encodes both image and action text descriptions using pre-trained vision-language models, aligning visual and linguistic representations in a unified semantic space. Unlike conventional cross-modal alignment methods, ACTOR employs language-guided attention to deeply model interaction semantics in the aligned space, enabling the model to infer fine-grained actions and produce semantically meaningful query representations. 
% To further address the issue of semantically ambiguous query initialization, 
In addition, we introduce the Perceptual Distilled Query Decoder (PDQD), which learns object-aware semantic features by distilling object category awareness from a pre-trained detector. PDQD incorporates high-quality object detection results and uses auxiliary supervision during training to guide the decoder in learning object categories and spatial cues more effectively. During inference, this dual-branch query initialization strategy—combining ACTOR's action-aware queries with PDQD's object-aware queries—provides semantically grounded and structurally informative representations, enabling more accurate and stable prediction of interaction triplets.
Main Contributions:

\begin{itemize}
% Here are the compressed contributions in passive voice:

    \item A novel Action-aware Cross-modal TransfORmer (ACTOR) is proposed, which leverages language-guided attention to infer interaction semantics and generate semantically meaningful query representations.
    
    \item A Perceptual Distilled Query Decoder (PDQD) is introduced to address semantic ambiguity in query initialization by distilling object category awareness from pre-trained detectors.
    
    \item A dual-branch query initialization strategy is designed that combines action-aware and object-aware queries for more effective HOI detection.
    
    \item State-of-the-art performance is achieved on HICO-Det and V-COCO benchmarks, demonstrating the effectiveness of incorporating semantic priors into HOI detection.
    % \item We propose a novel Action-aware Cross-modal TransfORmer that goes beyond simple modality alignment by leveraging language-guided attention to infer interaction semantics and generate semantically meaningful query representations for HOI detection.
    % \item We introduce a Perceptual Distilled Query Decoder that addresses the semantic ambiguity in query initialization by distilling object category awareness from pre-trained detectors, providing high-quality object queries with enhanced spatial and categorical understanding.
    % \item Dual-branch Query Initialization: We design a complementary query initialization strategy that combines action-aware queries from ACTOR with object-aware queries from PDQD, enabling more interpretable and effective query representations for detecting human-object interactions.
    % \item State-of-the-art Performance: Extensive experiments on HICO-Det and V-COCO benchmarks demonstrate that QueryCraft achieves superior performance and strong generalization capabilities, validating the effectiveness of incorporating semantic priors and guided feature learning into HOI detection.
\end{itemize}

\section{Related Work}

% \subsection{Human-Object Interaction Detection}
\subsection{Two-stage HOI Approaches.} 
% The early HOI models primarily used a two-stage process, where the instance was first detected and then action classification was performed. 
Early approaches in HOI detection adopted a sequential methodology, initially employing established object detection architectures~\cite{ren2015faster,girshick2015fast} to identify human and object instances, followed by a secondary phase that analyzed potential human-object combinations to determine interaction categories~\cite{chao2018learning,gao2018ican,kim2020detecting,zhang2021spatially,xu2022effective}. Notable examples include the HO-RCNN framework~\cite{chao2018learning} and the iCAN architecture~\cite{gao2018ican}. The latter introduced an instance-focused attention mechanism using visual features to highlight relevant regions for each detected entity. However, methods relying solely on visual features of isolated instances miss broader scene context, limiting interaction recognition. To overcome this, Wang et al. proposed an enhanced attention mechanism that integrates contextual understanding with appearance-based learning~\cite{wang2019deep}. Later, EfHOI~\cite{xu2022effective} introduced a human-centric reasoning architecture that uses contextual relationships to improve interaction prediction accuracy.

\subsection{Transformer-based HOI Detection}

% \textbf{One-stage methods.} 
% \textbf{One-stage methods.} 
Unlike sequential approaches, one-stage architectures (transformer-based) enable direct image-to-triplet transformation~\cite{chen2021reformulating,liao2022gen,peng2023parallel,wang2024ted,xue2025towards}. PPDM~\cite{liao2020ppdm} introduced geometric centerpoints between human-object pairs as interaction anchors, improving efficiency and accuracy. Though IP-Net~\cite{wang2020learning} used similar centerpoint strategies, it showed limited generalization~\cite{zou2021end}. 
Inspired by the success of DETR~\cite{carion2020end} in object detection, recent works have explored Transformer-based architectures for end-to-end HOI detection~\cite{zou2021end,liao2022gen,xue2025towards}, utilizing query embeddings for feature extraction.
% This led to transformer-based solutions like HOITrans~\cite{zou2021end} and QPIC~\cite{tamura2021qpic}, utilizing query embeddings for feature extraction. 
GEN-VLKT~\cite{liao2022gen} addressed post-processing overhead through guided embeddings and integrated CLIP~\cite{radford2021learning} for enhanced semantic understanding. For non-contact interactions, Wang et al.~\cite{wang2024ted} developed distributed decoding to capture contextual information around instances. KI2HOI~\cite{xue2025towards} improves HOI detection by integrating visual-language model knowledge, addressing the limitation of existing methods that rely on extensive manual annotations. However, existing DETR-based HOI methods use random queries to query for humans, objects, and interactions, which makes it difficult to learn query vectors for the challenging HOI task, affecting their overall performance~\cite{wang2024review}. 

\begin{figure*}[!ht]
\centering %表示居中
\includegraphics[width=\linewidth]{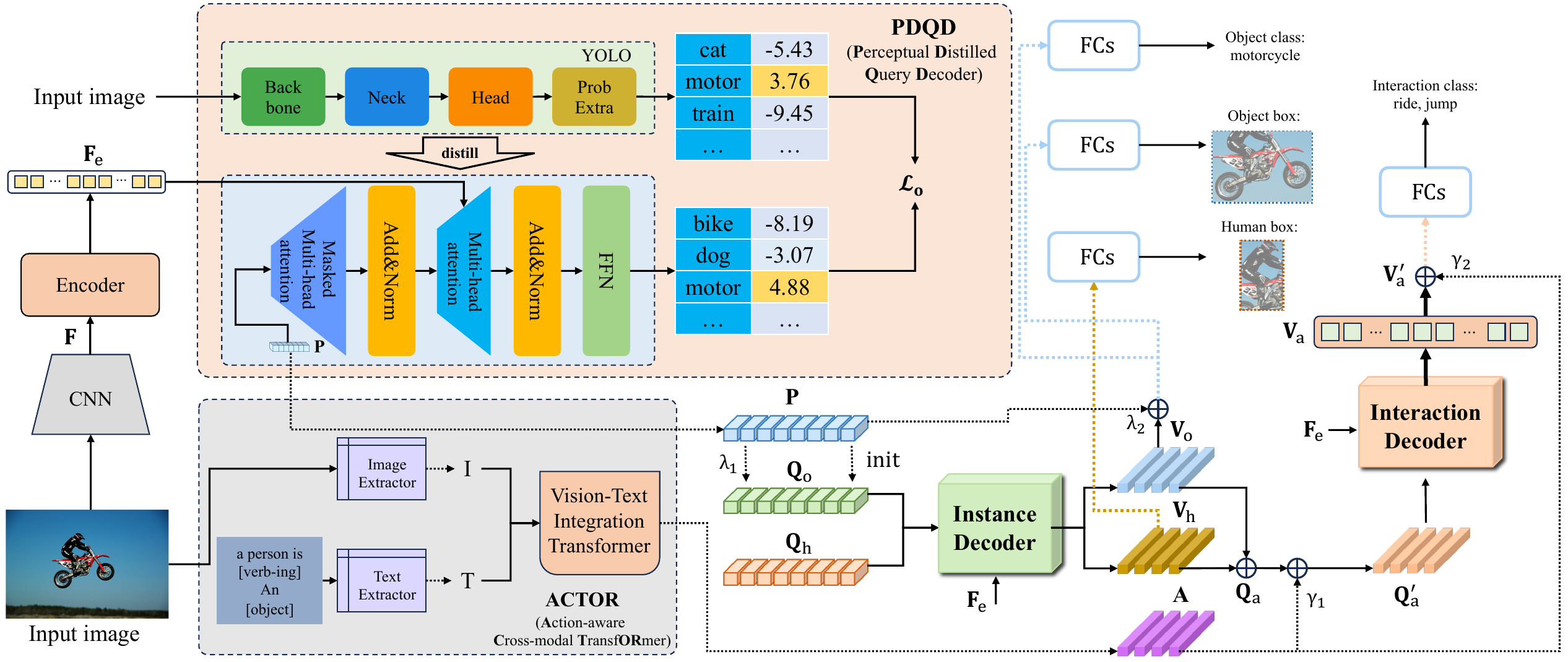}
% [height=4.5cm]表示高度
%[width=9.5cm]表示宽度
%{111.eps}表示eps格式的图片，名为111
\vspace{-0.5cm}
\caption{Detailed architecture of QueryCraft. The framework consists of three main components: (1) PDQD module (top) distills object knowledge from pre-trained YOLO to produce object queries $\mathbf{P}$ with semantic awareness; (2) ACTOR module (bottom left) extracts action-aware features by integrating visual and textual information through a Vision-Text Integration Transformer, generating interaction queries $\mathbf{A}$;  (3) Instance and Interaction Decoders (right) process these queries to predict human boxes ($\bm{V}_h$), object boxes/classes ($\bm{V}_o$), and interaction classifications ($\bm{V}_a$). The model jointly optimizes predictions through fully connected layers (FCs) to output object classes, bounding boxes, human boxes, and interaction classes.
}
%图片的名称
\label{fig:network}
%图片的标签，用于文章中的引用，注意到标签的数字与实际文章显示的数字可能不同
% \vspace{-0.2cm}
\end{figure*}

\section{Method}
% \subsection{Overview}

The overall architecture of QueryCraft is shown in Figure~\ref{fig:network}. Designed as a plug-and-play module, QueryCraft can be integrated into any query-based HOI detection framework. Its core contribution lies in two complementary modules that inject semantic priors into query initialization. The Perceptual Distilled Query Decoder (PDQD) learns object-aware representations via an auxiliary multi-label classification task. These representations not only initialize instance decoder queries but are also added to decoder outputs to improve object localization and object classification. In parallel, the Action-aware Cross-modal TransfORmer (ACTOR) exploits language priors by performing cross-modal attention between visual features and textual action prompts, producing action-aware queries that enhance interaction decoder outputs and improve action classification.

\subsection{HOI Pipeline}
% \subsection{Backbone and Feature Encoding}
% backbone用于特征提取。在HOI模型中，
Given an input image $\mathbf{X} \in \mathbb{R}^{3 \times H \times W}$, a backbone (e.g., ResNet-50~\cite{he2016deep} or Swin Transformer~\cite{liu2021swin}) is used to extract feature maps $\mathbf{F} \in \mathbb{R}^{C \times H' \times W'}$, where $H$ and $W$ denote the height and width of the input image, and $H'$ and $W'$ represent the height and width of the extracted feature map, respectively. $C$ denotes the number of feature channels. The input feature $\mathbf{F}$ is fed into the encoder network to extract global image features, denoted as $\mathbf{F}_e$. The instance decoder takes as input the encoded features $\mathbf{F}_e$ and the query matrices for humans and objects, denoted as $\mathbf{Q}_h \in \mathbb{R}^{N_q \times C'}$ and $\mathbf{Q}_o \in \mathbb{R}^{N_q \times C'}$, respectively, where $N_q$ is the number of queries and $C'$ is the feature dimension. It outputs human features $\mathbf{V}_h \in \mathbb{R}^{N_q \times C'}$ and object features $\mathbf{V}_o \in \mathbb{R}^{N_q \times C'}$. After passing through a fully connected layers (FCS), $\mathbf{V}_h$ and $\mathbf{V}_o$ produce the human bounding boxes $\mathbf{B}_h \in \mathbb{R}^{N_q \times 4}$, object bounding boxes $\mathbf{B}_o \in \mathbb{R}^{N_q \times 4}$, and object classes $\mathbf{O}_c \in \mathbb{R}^{N_q}$. Each 4-dimensional bounding box is represented as $(x_1, y_1, x_2, y_2)$, where $(x_1, y_1)$ denotes the top-left corner and $(x_2, y_2)$ denotes the bottom-right corner. Additionally, $\mathbf{V}_h$ and $\mathbf{V}_o$ are used as query matrices $\mathbf{Q}_a \in \mathbb{R}^{N_q \times C'}$ for the action decoder, which, together with $\mathbf{F}_e$, outputs action features $\mathbf{V}_a \in \mathbb{R}^{N_q \times C'}$. $\mathbf{V}_a$ are passed through an FCS layer to predict the final action classes $\mathbf{C} \in \mathbb{R}^{N_q}$. Ultimately, we obtain a set of quadruples $\mathbf{\mathbb{O}} = \{ \langle \mathbf{B}_h, \mathbf{B}_o, \mathbf{O}_c, \mathbf{C} \rangle \}$, each representing a human-object interaction. The total number of HOI predictions is $N_q$. A post-processing step, such as Non-Maximum Suppression (NMS), is applied to filter out redundant interaction pairs. However, due to the use of randomly initialized query matrices, the above pipeline suffers from ambiguous query semantics, which negatively impacts the overall detection performance.

\subsection{Perceptual Distilled Query Decoder (PDQD)}

The PDQD module addresses the limitation of random query initialization in DETR-based HOI detectors by learning object-aware query representations through an auxiliary multi-label classification task. This module generates semantically meaningful queries that encode object category information, which subsequently enhances both object detection and classification performance.

\subsubsection{Architecture}
The PDQD module includes a learnable projection tokens $\mathbf{P} \in \mathbb{R}^{N_q \times C'}$. $\mathbf{P}$ is processed through a transformer decoder to interact with the encoded image features:
\begin{equation}
\mathbf{F}_{obj} = \text{TransformerDecoder}(\mathbf{P}, \mathbf{F}_e),
\end{equation}
where $\mathbf{F}_e \in \mathbb{R}^{C \times H' \times W'}$ is the feature from the backbone encoder, and $\mathbf{F}_{obj} \in \mathbb{R}^{N_q \times C'}$ represents the object feature.

To enforce tht ability of object category awareness, we apply global average pooling followed by a multi-label classification head:
\begin{equation}
\mathbf{f}_{pool} = \frac{1}{N_q}\sum_{i=1}^{N_q} \mathbf{F}_{obj}^{(i)}, \quad\mathbf{f}_{pool} \in \mathbb{R}^{C'},
\end{equation}
\begin{equation}
\mathbf{y}^{cls} = \text{MLP}(\mathbf{f}_{pool}), \quad \mathbf{y}^{cls} \in \mathbb{R}^{N_{cls}},
\end{equation}
where $N_{cls}=80$ is the number of object categories, and the MLP consists of two linear layers with ReLU activation: $\text{Linear}(C' \rightarrow 128) \rightarrow \text{ReLU} \rightarrow \text{Linear}(128 \rightarrow N_{cls})$.

\subsubsection{Training Objective}
The PDQD module is trained to perform multi-label object classification by distilling knowledge from a pre-trained YOLO detector. For each training image, we first obtain YOLO's detection results and construct a pseudo ground-truth multi-label vector $\mathbf{y}^{L} \in \{0,1\}^{N_{cls}}$, where $y_j^{L} = 1$ if YOLO detects at least one instance of class $j$ with confidence score above threshold $\tau$ (we set $\tau=0.5$ in our experiments).

The PDQD module is then trained with a multi-label binary cross-entropy loss using these YOLO-generated labels:
\begin{equation}
\small
\mathcal{L}_{o} = -\frac{1}{N_{cls}}\sum_{j=1}^{N_{cls}} \left[ y_{j}^{L}\log(\sigma(y^{cls}_{j})) + (1-y_{j}^{L})\log(1-\sigma(y^{cls}_{j})) \right],
\end{equation}
where $y_{j}^{L} \in \{0,1\}$ indicates whether YOLO detected object class $j$ in image $i$, $y^{cls}_{j}$ is the logit predicted by our classification head, and $\sigma(\cdot)$ denotes the sigmoid function.
% This training strategy effectively transfers YOLO's robust object detection capabilities into our learnable projection tokens $\mathbf{P}$. By learning to predict the same object categories as YOLO, the projection tokens acquire rich object-centric representations that serve as strong semantic priors for query initialization. This approach is more reliable than using manual annotations, as YOLO's predictions inherently capture the visual detectability of objects, making the learned representations more suitable for the downstream detection task.

Through this design, the projection tokens $\mathbf{P}$ learn to encode rich object category information from the training data, effectively distilling perceptual knowledge that benefits both the initialization and final prediction stages of HOI detection.

\subsubsection{Query Enhancement}
The learned projection tokens $\mathbf{P}$ serve a dual purpose in our framework. First, they initialize the object queries in the instance decoder:
\begin{equation}
\mathbf{Q}_{o}^{'} = \mathbf{Q}_{o} + \lambda_1 \cdot \mathbf{P},
\end{equation}
where $\mathbf{Q}_{o}$ represents the original random initialization, and $\lambda_1$ is a weighting factor. We specifically enhance $\mathbf{Q}_o$ rather than $\mathbf{Q}_h$ because $\mathbf{Q}_h$ is solely responsible for human localization, while $\mathbf{Q}_o$ handles both localization and recognition across all object categories. Through knowledge distillation, $\mathbf{P}$ encapsulates rich object-centric information from the pre-trained detector, making it particularly suitable for initializing $\mathbf{Q}_o$ to improve multi-class object detection performance.

% The learned projection tokens $\mathbf{P}$ serve a dual purpose in our framework. First, they initialize the object queries in the instance decoder (我们不去增强Q_h是因为Q_h只会单独负责人体的定位，而Q_o负责所有物体类别的定位和识别,通过知识蒸馏后的P蕴含了物体的各种信息，因此适合初始化Q_o):

Second, after the instance decoder produces its output features $\mathbf{V}_{o} \in \mathbb{R}^{N_q \times C'}$, we further enhance them with the learned object representations:
\begin{equation}
\mathbf{V}_{o}^{'} = \mathbf{V}_{o} + \lambda_2 \cdot \mathbf{P}.
\end{equation}

The $\mathbf{V}_{o}^{'}$ is then used for predicting object bounding boxes, and object categories.

% \subsubsection{Training Objective}
% The PDQD module is trained with a multi-label binary cross-entropy loss:

% \begin{equation}
% \mathcal{L}_{pdqd} = -\frac{1}{B \cdot N_{cls}}\sum_{i=1}^{B}\sum_{j=1}^ge{N_{cls}} \left[ y_{ij}\log(\sigma(y^{cls}_{ij})) + (1-y_{ij})\log(1-\sigma(y^{cls}_{ij})) \right]
% \end{equation}

% where $y_{ij} \in \{0,1\}$ indicates whether object class $j$ is present in image $i$, $y^{cls}_{ij}$ is the predicted logit, and $\sigma(\cdot)$ denotes the sigmoid function.

% Through this design, the projection tokens $\mathbf{P}$ learn to encode rich object category information from the training data, effectively distilling perceptual knowledge that benefits both the initialization and final prediction stages of HOI detection.

\subsection{Action-aware Cross-modal Transformer (ACTOR)}
While PDQD provides object-aware initialization, accurate HOI detection also requires understanding the semantic space of possible interactions. The ACTOR module addresses this by leveraging language priors to generate action-aware query representations through cross-modal attention between visual and textual features.

\subsubsection{Motivation and Design}
HOI are inherently tied to semantic concepts expressed in natural language. For instance, the visual pattern of "riding a bicycle" corresponds to specific spatial relationships that can be described textually. ACTOR exploits this correspondence by using textual descriptions of all possible interaction categories as a semantic anchor space to guide query initialization. 

% For the linguistic branch, we construct a comprehensive action vocabulary by generating textual descriptions for all $N_{act}$ interaction categories using structured templates (e.g., "a person [action] a [object]"). These descriptions are encoded using the CLIP text encoder:
% Given a set of $N_{T}$ interaction categories (e.g., 600 for HICO-DET), we construct textual prompts for each action using templates $\mathcal{T}$ such as ``a person [action] a [object]''. These prompts are encoded using a pre-trained text encoder (e.g., CLIP text encoder) to obtain textual embeddings $\mathbf{T} \in \mathbb{R}^{N_{T} \times d_{t}}$:

Given a set of $N_{T}$ interaction categories (e.g., $\langle$ride, bicycle$\rangle$, $\langle$eat. banana$\rangle$, and so on.), we construct textual prompts for each action using templates $\mathcal{T}$. These prompts are encoded using a pre-trained text extractor to obtain textual embeddings $\mathbf{T} \in \mathbb{R}^{N_{T} \times d_{t}}$:
\begin{equation}
\mathbf{T} = \text{TextExtractor}(\mathcal{T}(v_c, o_c)),
\end{equation}
where $\mathcal{T}$ represents text template such as ``a person [action] a [object]'', $v_c$ denotes action, and $o_c$ denotes object. In addition, we utilize an image extractor to extract high-level visual features $\mathbf{I} \in \mathbb{R}^{1 \times d_{t}}$. 
% This yields a semantic dictionary $\mathbf{D} = \{\mathbf{t}_1, ..., \mathbf{t}_{N_{act}}\} \subset \mathcal{L}$. Crucially, this dictionary encodes not just individual actions but their compositional relationships with objects, providing a structured prior for interaction understanding.

% \subsubsection{Theoretical Foundation}
We hypothesize that the visual manifestation of HOI can be effectively characterized through their linguistic descriptions. Formally, let $\mathbf{I}$ denote the visual feature space and $\mathbf{T}$ the linguistic embedding space. We seek to learn a mapping $\phi: (\mathbf{I}, \mathbf{T}) \rightarrow \mathbf{A}$ where $\mathbf{A}$ represents a shared semantic space that captures the underlying interaction patterns. Our key insight is that pre-trained vision-language models (e.g., CLIP~\cite{radford2021learning} or BLIP~\cite{li2022blip,li2023blip}) have already learned rich alignments between $\mathbf{I}$ and $\mathbf{T}$. ACTOR exploits this alignment by constructing a semantic dictionary of interaction prototypes in the linguistic space, which serves as anchors for grounding visual queries.

% \subsubsection{Semantic Dictionary Construction}
% For each interaction category $c \in \{1, ..., N_{act}\}$, we generate structured linguistic representations using a template function $\mathcal{T}$:

% \begin{equation}
% \mathbf{t}_c = \text{TextExtractor}(\mathcal{T}(v_c, o_c))
% \end{equation}

% where $v_c$ denotes the verb (action) and $o_c$ the object category. This yields a semantic dictionary $\mathbf{D} = \{\mathbf{t}_1, ..., \mathbf{t}_{N_{act}}\} \subset \mathcal{L}$. Crucially, this dictionary encodes not just individual actions but their compositional relationships with objects, providing a structured prior for interaction understanding.

\subsubsection{Cross-modal Semantic Alignment}
ACTOR employs a novel asymmetric cross-attention mechanism where visual features query the semantic dictionary to discover relevant interaction patterns. $\mathbf{I}$ is expand it into $N_q$ query seeds:
\begin{equation}
\mathbf{Q}^{(0)} = \mathbf{R}^\top\mathbf{I},
\end{equation}
% where $\mathbf{I} \in \mathbb{R}^{1 \times N}$, $\mathbf{R} \in \mathbb{R}^{1 \times N_q}$, and $\mathbf{M} \in \mathbb{R}^{N_q \times N}$.
where $\mathbf{I} \in \mathbb{R}^{1 \times N}$, $\mathbf{R} \in \mathbb{R}^{1 \times N_q}$ (with $r_i = 1, \forall i, r_i \in \mathbf{R}$), and $\mathbf{Q}^{(0)} \in \mathbb{R}^{N_q \times N}$.
% where $\otimes$ denotes the expansion operation that replicates the visual feature $N_q$ times.
% Given the encoded visual memory $\mathbf{M}$, we first compute a global visual representation:
% \begin{equation}
% \mathbf{g} = \psi(\mathbf{M}) = \frac{1}{|\mathbf{M}|}\sum_{\mathbf{m} \in \mathbf{M}} f_{\theta}(\mathbf{m})
% \end{equation}
% where $f_{\theta}$ is a learned transformation. This global representation is then expanded into $N_q$ query seeds through a learnable expansion function:
% \begin{equation}
% \mathbf{Q}^{(0)} = \text{Expand}(\mathbf{g}) = \mathbf{g} \mathbf{W}_{\text{expand}} + \mathbf{b}_{\text{expand}}
% \end{equation}
The cross-modal refinement then processes these features through $L$ layers (we use $L=3$), where each layer $l \in \{1, ..., L\}$  consists of: 
% The core innovation lies in our multi-layer cross-modal refinement process:
\begin{equation}
\mathbf{Q}^{(l+1)} = \mathcal{F}_{\text{ACTOR}}(\mathbf{Q}^{(l)}, \mathbf{T}),
\end{equation}
where $\mathcal{F}_{\text{ACTOR}}$ implements a specialized transformer block that performs:
\begin{equation}
\begin{aligned}
\mathbf{S}^{(l)} &= \mathcal{S}\left(\frac{\mathbf{Q}^{(l)}\mathbf{W}_Q (\mathbf{T}\mathbf{W}_K)^\top}{\sqrt{d_t}}\right), \\
\tilde{\mathbf{Q}}^{(l)} &= \mathbf{S}^{(l)}(\mathbf{T}\mathbf{W}_V), \\
\mathbf{Q}^{(l+1)} &= \text{FFN}(\text{LN}(\mathbf{Q}^{(l)} + \tilde{\mathbf{Q}}^{(l)})).
\end{aligned}
\end{equation}

The attention weights $\mathbf{S}^{(l)}$ can be interpreted as a soft assignment of visual queries to semantic interaction prototypes, effectively performing implicit reasoning about which interactions are plausible given the visual context.

\begin{table*}[ht]
\centering
\setlength{\tabcolsep}{1.5mm}{
\begin{tabular}{lllccc}
\specialrule{1.5pt}{0pt}{0pt}
\multicolumn{1}{l|}{\multirow{2}{*}{Method}} & \multicolumn{1}{l|}{\multirow{2}{*}{Backbone}} & \multicolumn{1}{l|}{\multirow{2}{*}{Detector}} & \multicolumn{3}{c}{\textbf{mAP}$\uparrow$}  \\ \cline{4-6} 
\multicolumn{1}{c|}{}                        & \multicolumn{1}{c|}{}                          & \multicolumn{1}{c|}{}                          & Full  & Rare  & Non-Rare \\ \hline
% \multicolumn{6}{c}{Fully Supervised (using \textcolor[RGB]{210,0,0}{$\langle$``human'', ``action'', ``object''$\rangle$} labels) \textbf{Task Level: Easy}}                                                                                                                                      \\
\multicolumn{1}{l|}{HO-RCNN~\cite{chao2018learning}}           & \multicolumn{1}{l|}{CaffeNet}  & \multicolumn{1}{l|}{Fast R-CNN}         & 7.81                     & 5.37                     & 8.54                        \\
\multicolumn{1}{l|}{InteractNet~\cite{gkioxari2018detecting}}           & \multicolumn{1}{l|}{ResNet-50}             & \multicolumn{1}{l|}{Faster R-CNN}                 & 9.94                     & 7.16                     & \multicolumn{1}{c}{10.77}     \\
\multicolumn{1}{l|}{iCAN~\cite{gao2018ican}}                    & \multicolumn{1}{l|}{ResNet-50}                  & \multicolumn{1}{l|}{Faster R-CNN}       & 14.84 & 10.45 & 16.15    \\
\multicolumn{1}{l|}{UnionDet~\cite{kim2020uniondet}}          & \multicolumn{1}{l|}{ResNet-50-FPN}     & \multicolumn{1}{l|}{RetinaNet}       & 17.58   & 11.72  & 19.33     \\

\multicolumn{1}{l|}{IP-Net~\cite{wang2020learning}}           & \multicolumn{1}{l|}{Hourglass-104}     & \multicolumn{1}{l|}{FPN}      & 19.56   & 12.79  & 21.58        \\

\multicolumn{1}{l|}{VSGNet~\cite{ulutan2020vsgnet}}                  & \multicolumn{1}{l|}{ResNet-152}                 & \multicolumn{1}{l|}{Faster R-CNN}                          & 19.80 & 16.05 & 20.91    \\

\multicolumn{1}{l|}{FCMNet~\cite{liu2020amplifying}}                          & \multicolumn{1}{l|}{ResNet-50}             & \multicolumn{1}{l|}{Faster R-CNN}                 & 20.41                       & 17.34                   & 21.56     \\
\multicolumn{1}{l|}{ACP~\cite{kim2020detecting}}              & \multicolumn{1}{l|}{Res-DCN-152}     & \multicolumn{1}{l|}{Faster R-CNN}    & 20.59   & 15.92  & 21.98     \\

\multicolumn{1}{l|}{PD-Net~\cite{zhong2020polysemy}}          & \multicolumn{1}{l|}{ResNet-152-FPN}     & \multicolumn{1}{l|}{Faster R-CNN} & 20.81   & 15.90  & 22.28    \\
\multicolumn{1}{l|}{SG2HOI~\cite{he2021exploiting}}           & \multicolumn{1}{l|}{ResNet-50}     & \multicolumn{1}{l|}{Faster R-CNN}      & 20.93   & 18.24  & 21.78      \\
\multicolumn{1}{l|}{DJ-RN~\cite{li2020detailed}}              & \multicolumn{1}{l|}{ResNet-50}    & \multicolumn{1}{l|}{Faster R-CNN}      & 21.34   & 18.53  & 22.18        \\

\multicolumn{1}{l|}{SCG~\cite{zhang2021spatially}}                     & \multicolumn{1}{l|}{ResNet-50}              & \multicolumn{1}{l|}{Faster R-CNN}   & 21.85 & 18.11 & 22.97    \\
% \multicolumn{1}{l|}{IDN~\cite{li2020hoi}}                    & \multicolumn{1}{l|}{ResNet-50}                  & \multicolumn{1}{l|}{Faster R-CNN}   & 23.36 & 22.47 & 23.63    \\
\multicolumn{1}{l|}{HOTR~\cite{kim2021hotr}}                    & \multicolumn{1}{l|}{ResNet-50}                  & \multicolumn{1}{l|}{DETR}           & 25.10 & 17.34 & 27.42    \\
\multicolumn{1}{l|}{MSTR~\cite{kim2022mstr}}                    & \multicolumn{1}{l|}{ResNet-50}                  & \multicolumn{1}{l|}{DETR}           & 31.17 & 25.31 & 33.92    \\
% \multicolumn{1}{l|}{MSTR~\cite{kim2022mstr}}                    & \multicolumn{1}{l|}{ResNet-50}                  & \multicolumn{1}{l|}{DETR}           & 32.64 & 28.90 & 33.76    \\
% \multicolumn{1}{l|}{CycleHOI~\cite{wang2024cyclehoi}}                    & \multicolumn{1}{l|}{ResNet-50}                  & \multicolumn{1}{l|}{DETR}           & 32.23 & 25.27 & 34.01    \\
% \multicolumn{1}{l|}{GEN-VLKT~\cite{liao2022gen}}                 & \multicolumn{1}{l|}{ResNet-50}                  & \multicolumn{1}{l|}{DETR}               & 33.75      & 29.25      & 35.10         \\
% \multicolumn{1}{l|}{MSTR~\cite{kim2022mstr}}                    & \multicolumn{1}{l|}{ResNet-50}                  & \multicolumn{1}{l|}{DETR}           & 31.17 & 25.31 & 33.92    \\
% \multicolumn{1}{l|}{PViC~\cite{zhang2023exploring}}                    & \multicolumn{1}{l|}{ResNet-50}                  & \multicolumn{1}{l|}{DETR}           & 34.69 & 32.14 & 35.45    \\
\rowcolor[gray]{0.95}\multicolumn{1}{l|}{GEN-VLKT~\cite{liao2022gen}}                 & \multicolumn{1}{l|}{ResNet-50}                  & \multicolumn{1}{l|}{DETR}               & 33.51      & 28.52      & 35.00         \\
\rowcolor[gray]{0.9}\multicolumn{1}{l|}{RLIPv2~\cite{yuan2023rlipv2}}                    & \multicolumn{1}{l|}{ResNet-50}                  & \multicolumn{1}{l|}{DETR}           & 35.38 & 29.61 & 37.10 \\
\rowcolor[gray]{0.85}\multicolumn{1}{l|}{RLIPv2~\cite{yuan2023rlipv2}}                    & \multicolumn{1}{l|}{Swin-L}                  & \multicolumn{1}{l|}{DETR}           & 45.09 & 43.23 & 45.64 \\
\rowcolor[gray]{0.8}\multicolumn{1}{l|}{TED-Net~\cite{wang2024ted}}                    & \multicolumn{1}{l|}{ResNet-50}                  & \multicolumn{1}{l|}{DETR}           & 34.00 & 29.88 & 35.24    \\
\rowcolor[gray]{0.75}\multicolumn{1}{l|}{LOGICHOI~\cite{li2024neural}}                    & \multicolumn{1}{l|}{ResNet-50}                  & \multicolumn{1}{l|}{DETR}           & 35.47 & 32.03 & 36.22   \\
\rowcolor[gray]{0.7}\multicolumn{1}{l|}{KI2HOI~\cite{xue2025towards}}                    & \multicolumn{1}{l|}{ResNet-50}                  & \multicolumn{1}{l|}{DETR}           & 34.20 & 32.26 & 36.10    \\
\hline \hline
\rowcolor[gray]{0.95}\multicolumn{1}{l|}{GEN-VLKT\textbf{ + }Ours}         & \multicolumn{1}{l|}{ResNet-50}               & \multicolumn{1}{l|}{DETR}                          & \begin{tabular}[c]{@{}c@{}}34.63\\ \small+\textbf{1.12}\end{tabular} & \begin{tabular}[c]{@{}c@{}}30.24\\ \small+\textbf{1.72}\end{tabular}  & \begin{tabular}[c]{@{}c@{}}35.95\\ \small+\textbf{0.95}\end{tabular}    \\
\rowcolor[gray]{0.9}\multicolumn{1}{l|}{RLIPv2\textbf{ + }Ours}                    & \multicolumn{1}{l|}{ResNet-50}                  & \multicolumn{1}{l|}{DETR}           & \begin{tabular}[c]{@{}c@{}}36.42\\ \small+\textbf{1.04}\end{tabular} & \begin{tabular}[c]{@{}c@{}}30.84\\ \small+\textbf{1.23}\end{tabular} & \begin{tabular}[c]{@{}c@{}}38.01\\ \small+\textbf{0.91}\end{tabular} \\
\rowcolor[gray]{0.85}\multicolumn{1}{l|}{RLIPv2\textbf{ + }Ours}                    & \multicolumn{1}{l|}{Swin-L}                  & \multicolumn{1}{l|}{DETR}           & \begin{tabular}[c]{@{}c@{}}46.01\\ \small+\textbf{0.92}\end{tabular} & \begin{tabular}[c]{@{}c@{}}44.75\\ \small+\textbf{1.52}\end{tabular} & \begin{tabular}[c]{@{}c@{}}46.42\\ \small+\textbf{0.78}\end{tabular} \\
\rowcolor[gray]{0.8}\multicolumn{1}{l|}{TED-Net\textbf{ + }Ours}                  & \multicolumn{1}{l|}{ResNet-50}                 & \multicolumn{1}{l|}{DETR}                          & \begin{tabular}[c]{@{}c@{}}35.17\\ \small+\textbf{1.17}\end{tabular} & \begin{tabular}[c]{@{}c@{}}31.29\\ \small+\textbf{1.41}\end{tabular} & \begin{tabular}[c]{@{}c@{}}36.16\\ \small+\textbf{0.92}\end{tabular}    \\
\rowcolor[gray]{0.75}\multicolumn{1}{l|}{LOGICHOI\textbf{ + }Ours}                    & \multicolumn{1}{l|}{ResNet-50}                  & \multicolumn{1}{l|}{DETR}           & \begin{tabular}[c]{@{}c@{}}36.48\\ \small+\textbf{1.01}\end{tabular} & \begin{tabular}[c]{@{}c@{}}33.24\\ \small+\textbf{1.21}\end{tabular} & \begin{tabular}[c]{@{}c@{}}37.16\\ \small+\textbf{0.94}\end{tabular}   \\
\rowcolor[gray]{0.7}\multicolumn{1}{l|}{KI2HOI\textbf{ + }Ours}                 & \multicolumn{1}{l|}{ResNet-50}                  & \multicolumn{1}{l|}{DETR}                          & \begin{tabular}[c]{@{}c@{}}35.27\\ \small+\textbf{1.07}\end{tabular} & \begin{tabular}[c]{@{}c@{}}33.69\\ \small+\textbf{1.43}\end{tabular} & \begin{tabular}[c]{@{}c@{}}36.99\\ \small+\textbf{0.89}\end{tabular}    \\
\specialrule{1.5pt}{0pt}{0pt}
      
\end{tabular}

}
\caption{Performance comparisons on HICO-Det Dataset.}
\label{tab:hico_results}
\vspace{-0.3cm}
\end{table*}

\subsubsection{Action Query Generation}
The output of ACTOR, $\mathbf{A} = \mathbf{Q}^{(L)} \in \mathbb{R}^{N_q \times C'}$, represents action-aware queries that encode interaction semantics. These queries capture which actions are likely to occur in the image by attending to relevant textual descriptions. Similar to PDQD, these action-aware representations serve dual purposes:
\begin{equation}
\mathbf{Q}_{a}^{'} = \mathbf{Q}_{a} + \gamma_1 \cdot \mathbf{A}.
\end{equation}

Additionally, after the interaction decoder processes these initialized queries, we apply residual enhancement:
\begin{equation}
\mathbf{V}_{a}^{'} = \mathbf{V}_{a} + \gamma_2 \cdot \mathbf{A},
\end{equation}
where $\mathbf{Q}_{a}^{'}$ initializes the interaction decoder queries, and $\mathbf{V}_{a}^{'}$ enhances the interaction decoder outputs for improved action classification.

The cross-modal semantic alignment in ACTOR effectively creates a soft retrieval process over the action vocabulary. By learning to attend to relevant action descriptions based on visual content, the model develops a semantic understanding of potential interactions. Through this design, ACTOR provides action-specific semantic query initialization, which, in conjunction with PDQD, forms a comprehensive semantic prior, enabling effective HOI detection.

\section{Experiments}

\subsection{Datasets}
We evaluate QueryCraft on two HOI detection benchmarks: HICO-DET and V-COCO. HICO-DET contains 47,776 images (38,118 training, 9,658 testing) with 600 HOI categories formed by 80 objects and 117 action verbs. Performance is measured by mAP under Full (all 600 categories), Rare (138 categories with fewer than 10 training instances) settings, and Non-Rare (the remaining categories). V-COCO comprises 10,346 images with 80 interaction categories covering 29 action types. HICO-DET tests scalability with its large vocabulary and long-tail distribution, while V-COCO evaluates fine-grained action understanding through role-based annotations.

\subsection{Effectiveness for Regular HOI Detection}
\subsubsection{Results on HICO-DET}
Table~\ref{tab:hico_results} presents our results on HICO-DET. QueryCraft demonstrates consistent improvements across all baseline methods, with gains ranging from +0.92 to +1.17 mAP on the Full set. The improvements are most pronounced on the Rare subset, averaging +1.42 mAP across all methods. For instance, GEN-VLKT achieves +1.72 mAP on Rare interactions, while RLIPv2 with Swin-L backbone still gains +1.52 despite its strong baseline (43.23 mAP). These substantial improvements on rare categories validate that our semantic initialization provides crucial inductive biases for limited-data scenarios. The consistent gains across diverse architectures confirm that QueryCraft addresses a fundamental limitation in DETR-based HOI detection rather than being method-specific.
% 结果可视化见支撑材料。
\begin{table}[!h]
\centering

\setlength{\tabcolsep}{1.5mm}{
\small
\begin{tabular}{lllc}
\specialrule{1.5pt}{0pt}{0pt}
\multicolumn{1}{l|}{Method}      & \multicolumn{1}{l|}{Backbone}     & \multicolumn{1}{l|}{Detector}     & \textbf{mAP}$\uparrow$   \\ \hline
\multicolumn{1}{l|}{GEN-VLKT}        & \multicolumn{1}{l|}{ResNet50}     & \multicolumn{1}{l|}{DETR}         & 62.4 \\ 
\multicolumn{1}{l|}{RLIPv2}        & \multicolumn{1}{l|}{ResNet50}     & \multicolumn{1}{l|}{DETR}         & 65.9 \\ 
\multicolumn{1}{l|}{RLIPv2}        & \multicolumn{1}{l|}{Swin-L}     & \multicolumn{1}{l|}{DETR}         & 72.1 \\ 
\multicolumn{1}{l|}{TED-Net}        & \multicolumn{1}{l|}{ResNet50}     & \multicolumn{1}{l|}{DETR}         & 63.4 \\
\multicolumn{1}{l|}{LOGICHOI}        & \multicolumn{1}{l|}{ResNet50}     & \multicolumn{1}{l|}{DETR}         & 64.4 \\
\multicolumn{1}{l|}{KI2HOI}        & \multicolumn{1}{l|}{ResNet50}     & \multicolumn{1}{l|}{DETR}         & 63.9 \\\hline \hline

\multicolumn{1}{l|}{GEN-VLKT\textbf{ + }Ours}        & \multicolumn{1}{l|}{ResNet50}     & \multicolumn{1}{l|}{DETR}         & 63.7 (\small +\textbf{1.3})\\ 
\multicolumn{1}{l|}{RLIPv2\textbf{ + }Ours}        & \multicolumn{1}{l|}{ResNet50}     & \multicolumn{1}{l|}{DETR}         & 67.3 (\small +\textbf{1.4}) \\
\multicolumn{1}{l|}{RLIPv2\textbf{ + }Ours}        & \multicolumn{1}{l|}{Swin-L}     & \multicolumn{1}{l|}{DETR}         & 73.1 (\small +\textbf{1.0}) \\
\multicolumn{1}{l|}{TED-Net\textbf{ + }Ours}        & \multicolumn{1}{l|}{ResNet50}     & \multicolumn{1}{l|}{DETR}         & 65.1 (\small +\textbf{1.7})\\
\multicolumn{1}{l|}{LOGICHOI\textbf{ + }Ours}        & \multicolumn{1}{l|}{ResNet50}     & \multicolumn{1}{l|}{DETR}         & 65.6 (\small +\textbf{1.2})\\
\multicolumn{1}{l|}{K2HOI\textbf{ + }Ours}        & \multicolumn{1}{l|}{ResNet50}     & \multicolumn{1}{l|}{DETR}         & 65.0 (\small +\textbf{1.1})\\
\specialrule{1.5pt}{0pt}{0pt}

\end{tabular}}

\caption{Performance comparisons on V-COCO Dataset.}
\label{tab:vcoco_results}

\end{table}

\begin{table}[h]
    \centering
    \setlength{\tabcolsep}{0.3mm}{
    \small
    \begin{threeparttable}
    \begin{tabular}{ll|ccc}
\specialrule{1.5pt}{0pt}{0pt}
\multirow{2}{*}{Method}                                                                         & \multirow{2}{*}{Source}                               & \multicolumn{3}{c}{\textbf{mAP}$\uparrow$}                                                                                                                                                 \\
                                                                                                &                                                     & Full                                                    & Unseen                                                    & Seen                                                \\ \hline

GEN-VLKT & UV & 28.74 & 20.96 & 30.23 \\
RLIPv2  & UV & - & - & - \\
KI2HOI & UV & 31.85 & 25.20 & 32.95 \\
\hline \hline
GEN-VLKT\textbf{ + }Ours     & UV & 30.08 (\scriptsize +\textbf{1.34}) & 23.06 (\scriptsize +\textbf{2.10}) & 31.37 (\scriptsize +\textbf{1.14})  \\ 
KI2HOI\textbf{ + }Ours     & UV & 33.01 (\scriptsize +\textbf{1.16}) & 26.76 (\scriptsize +\textbf{1.56}) & 34.32 (\scriptsize +\textbf{1.37})  \\ \hline
GEN-VLKT  & UO & 25.63 & 10.51 & 28.92 \\
RLIPv2  & UO & - & - & - \\
KI2HOI & UO & 28.84 & 16.50 & 31.70 \\ \hline \hline
GEN-VLKT\textbf{ + }Ours     & UO & 27.21 (\scriptsize +\textbf{1.58}) & 12.96 (\scriptsize +\textbf{2.45}) & 30.65 (\scriptsize +\textbf{1.73})  \\ 
KI2HOI\textbf{ + }Ours     & UO & 30.11 (\scriptsize +\textbf{1.27}) & 18.37 (\scriptsize +\textbf{1.87}) & 32.94 (\scriptsize +\textbf{1.24})  \\ \hline
GEN-VLKT & NF-UC & 23.71 & 25.05 & 23.38 \\
RLIPv2 & NF-UC & 36.94 & 22.65 & 40.51 \\
KI2HOI & NF-UC & 27.77 & 28.89 & 28.31 \\
\hline \hline
GEN-VLKT\textbf{ + }Ours     & NF-UC & 24.69 (\scriptsize +\textbf{0.98}) & 26.25 (\scriptsize +\textbf{1.20}) & 24.27 (\scriptsize +\textbf{0.89})  \\ 
RLIPv2\textbf{ + }Ours     & NF-UC & 38.07 (\scriptsize +\textbf{1.13}) & 24.73 (\scriptsize +\textbf{2.08}) & 41.57 (\scriptsize +\textbf{1.06})  \\ 
KI2HOI\textbf{ + }Ours     & NF-UC & 28.92 (\scriptsize +\textbf{1.15}) & 30.13 (\scriptsize +\textbf{1.24}) & 29.42 (\scriptsize +\textbf{1.11})  \\ \hline
GEN-VLKT & RF-UC  & 30.56 & 21.36 & 32.91 \\
RLIPv2 & RF-UC & 42.26 & 31.23 & 45.01 \\
KI2HOI & RF-UC & 34.10 & 26.33 & 35.79 \\
\hline \hline
GEN-VLKT\textbf{ + }Ours     & RF-UC & 31.61 (\scriptsize +\textbf{1.05}) & 22.89 (\scriptsize +\textbf{1.53}) & 33.84 (\scriptsize +\textbf{0.93})  \\ 
RLIPv2\textbf{ + }Ours     & RF-UC & 43.48 (\scriptsize +\textbf{1.22}) & 33.02 (\scriptsize +\textbf{1.79}) & 45.94 (\scriptsize +\textbf{0.93})  \\ 
KI2HOI\textbf{ + }Ours     & RF-UC & 35.32 (\scriptsize +\textbf{1.22}) & 27.85 (\scriptsize +\textbf{1.52}) & 36.76 (\scriptsize +\textbf{0.97})  \\ \hline
 \specialrule{1.5pt}{0pt}{0pt}
\end{tabular}

\begin{tablenotes}
        \footnotesize
        \item[*] RLIPv2 uses Swin-L as backbone. GEN-VLKT and KI2HOI use ResNet-50 as backbone.
\end{tablenotes}
\end{threeparttable}

}
\caption{Performance comparison for zero-shot
HOI detection on HICO-Det.}
\label{tab:zero_shot}
% \vspace{-0.3cm}
\end{table}

\subsubsection{Results on V-COCO}
Table~\ref{tab:vcoco_results} shows our evaluation on V-COCO, where QueryCraft achieves improvements ranging from +1.0 mAP to +1.7 mAP points. TED-Net shows the largest gain (+1.7 mAP), and other methods demonstrate consistent improvements around +1.1 mAP to +1.4 mAP. Notably, even RLIPv2 with Swin-L, which already achieves 72.1 mAP, benefits from our approach (+1.0 mAP). The effectiveness on V-COCO's fine-grained action recognition task, which requires precise object localization and role understanding, further validates that both PDQD's object-aware and ACTOR's action-semantic initializations contribute meaningfully to interaction detection across different evaluation protocols.

\subsubsection{Zero-Shot Generalization}
Table~\ref{tab:zero_shot} presents zero-shot experiments across four protocols on HICO-DET. QueryCraft consistently enhances generalization capabilities across all settings. In Unseen Verb (UV), our method achieves substantial gains, particularly on unseen interactions (GEN-VLKT: +2.10, KI2HOI: +1.56 mAP), demonstrating ACTOR's effectiveness in reasoning about novel actions through language-guided attention. For Unseen Object (UO), improvements are even more pronounced, with GEN-VLKT gaining +2.45 mAP on unseen, showcasing PDQD's ability to transfer object knowledge to unseen categories. In compositional settings (NF-UC and RF-UC), QueryCraft maintains robust improvements, with RLIPv2 achieving +2.08 and +1.79 mAP on unseen subsets respectively. PDQD and ACTOR work synergistically—PDQD enables novel object recognition through distilled detection knowledge, while ACTOR facilitates understanding of new actions via cross-modal reasoning. This dual semantic grounding proves particularly valuable in zero-shot scenarios where random initialization lacks meaningful priors, fundamentally enhancing the model's ability to generalize beyond its training distribution.

% The consistent gains across all protocols (averaging +1.87 mAP on Rare) reveal that semantic query initialization provides crucial inductive biases for compositional reasoning. PDQD and ACTOR work synergistically—PDQD enables novel object recognition through distilled detection knowledge, while ACTOR facilitates understanding of new actions via cross-modal reasoning. This dual semantic grounding proves particularly valuable in zero-shot scenarios where random initialization lacks meaningful priors, fundamentally enhancing the model's ability to generalize beyond its training distribution.

\subsubsection{Intransitive Interaction Detection}
Table~\ref{tab:intransitive} presents results on intransitive (non-contact) interactions using HICO-DET-IC and V-COCO-IC benchmarks. On HICO-DET-IC, QueryCraft achieves substantial improvements, with GEN-VLKT gaining +1.84 mAP on Rare interactions and TED-Net showing consistent gains across all subsets (+1.33 mAP Full). Both methods achieve +1.22 mAP improvement on V-COCO-IC. The strong performance on intransitive interactions, which require understanding spatial relationships rather than physical contact, demonstrates the versatility of our approach. ACTOR's language-guided initialization proves particularly valuable as intransitive actions (e.g., ``watching'', ``flying'') are better characterized linguistically than visually. Combined with PDQD's accurate object localization for spatial reasoning, QueryCraft enables effective detection of both contact-based and abstract relationships, showcasing its applicability to comprehensive real-world scene understanding.

\begin{figure*}[ht]
\centering
% 第一行的两个表格
\begin{minipage}[t]{0.58\textwidth}
\centering
\setlength{\tabcolsep}{0.3mm}{%
% \small
\begin{tabular}{l|ccc|c}
\hline  \specialrule{1.5pt}{0pt}{0pt}
\rowcolor[HTML]{EFEFEF} 
 \cellcolor[HTML]{EFEFEF}                                            & \multicolumn{3}{c|}{\cellcolor[HTML]{EFEFEF}HICO-Det-IC} & \cellcolor[HTML]{EFEFEF}                         \\
\rowcolor[HTML]{EFEFEF} 
\multirow{-2}{*}{\cellcolor[HTML]{EFEFEF}Method}  & Full            & Rare            & Non-Rare          & \multirow{-2}{*}{\cellcolor[HTML]{EFEFEF}V-COCO-IC} \\ \hline

GEN-VLKT                & 27.25   & 23.73  & 28.03             & 33.59                        \\
TED-Net                 & 30.09         & 30.09        &  30.24           &  38.71                       \\ \hline \hline
GEN-VLKT\textbf{ + }Ours           &  28.45(\scriptsize +\textbf{1.12})       & 25.57(\scriptsize +\textbf{1.84})       &  29.13(\scriptsize +\textbf{1.09})          & 34.81(\scriptsize +\textbf{1.22})                      \\
TED-Net\textbf{ + }Ours            &  31.42(\scriptsize +\textbf{1.33})        &  31.22(\scriptsize +\textbf{1.13})       &  31.36(\scriptsize +\textbf{1.12})           &  39.93(\scriptsize +\textbf{1.22})                       \\ \hline  \specialrule{1.5pt}{0pt}{0pt}
\end{tabular}}
\captionof{table}{Performance comparisons on HICO-Det-IC and V-COCO-IC.}
\label{tab:intransitive}
\end{minipage}
\hfill
\begin{minipage}[t]{0.40\textwidth}
\centering
\vspace{-1.3cm}
% \vspace{+0.5mm}
\setlength{\tabcolsep}{0.35mm}{%
\begin{threeparttable}
\begin{tabular}{c|clcc}
\hline \specialrule{1.5pt}{0pt}{0pt}
\cellcolor[HTML]{EFEFEF}Method & \cellcolor[HTML]{EFEFEF}Epochs & \cellcolor[HTML]{EFEFEF} & \cellcolor[HTML]{EFEFEF}Epochs & \cellcolor[HTML]{EFEFEF}Reduction \\ \hline
GEN-VLKT                                & 87                             &                          & 76                             & $\downarrow$12.6\% \\
TED-Net                                 & 97                             &                          & 79                             & $\downarrow$18.6\% \\
LOGICHOI                                & 79                             &                          & 68                             & $\downarrow$13.9\% \\
KI2HOI                                  & 83                             & \multirow{-4}{*}{\centering\textbf{+Ours}}  & 72                             & $\downarrow$13.3\% \\ \hline \specialrule{1.5pt}{0pt}{0pt}
\end{tabular}%
\begin{tablenotes}
        \footnotesize
        \item[*] Best epoch numbers are obtained from our reproduction using official code implementations.
\end{tablenotes}
\end{threeparttable}

}
\vspace{-0.2cm}
% {\footnotesize Note: Best epoch numbers are obtained from our reproduction using official code implementations.}
\captionof{table}{Convergence epochs comparison.}
\label{tab:training_efficiency}
\end{minipage}

\vspace{0.2cm}

% 第二行的两个表格
\begin{minipage}[t]{0.48\textwidth}
\centering
\begin{tabular}{c|cccc|c}
\hline \specialrule{1.5pt}{0pt}{0pt}
\rowcolor[HTML]{EFEFEF} 
 \cellcolor[HTML]{EFEFEF}    &  \cellcolor[HTML]{EFEFEF}                                         & \multicolumn{3}{c|}{\cellcolor[HTML]{EFEFEF}HICO-Det} & \cellcolor[HTML]{EFEFEF}                         \\
\rowcolor[HTML]{EFEFEF} 
\multirow{-2}{*}{\cellcolor[HTML]{EFEFEF}ACTOR} & \multirow{-2}{*}{\cellcolor[HTML]{EFEFEF}PDQD} & Full            & Rare            & Non-Rare          & \multirow{-2}{*}{\cellcolor[HTML]{EFEFEF}V-COCO} \\ \hline

% \multirow{2}{*}{ACTOR} & \multirow{2}{*}{PDQD} & \multicolumn{3}{c}{HICO-Det} & \multirow{2}{*}{V-COCO} \\ 
                                           % &                                            & Full    & Rare    & Non-Rare  &                         \\ \hline
$\circ$     & $\circ$    &  33.51    &   28.52   &    35.00 & 62.4     \\
$\circ$     & $\bullet$  &   33.83   &    29.10  &      35.13 & 62.8    \\
$\bullet$   & $\circ$    &    34.24  &   29.91   &     35.58 & 63.1    \\
$\bullet$   & $\bullet$  &   \textbf{34.63}   &   \textbf{30.24}   &       \textbf{35.95} & \textbf{63.7}  \\ \hline \specialrule{1.5pt}{0pt}{0pt}
\end{tabular}
\captionof{table}{Performance of different components on the HICO-Det and V-COCO.}
\label{tab:ab_commpoent}
\end{minipage}
\hfill
\begin{minipage}[t]{0.48\textwidth}
\centering
\setlength{\tabcolsep}{1.5mm}{%
\begin{tabular}{ll|ccc|c}
\hline \specialrule{1.5pt}{0pt}{0pt}
\rowcolor[HTML]{EFEFEF} 
 \cellcolor[HTML]{EFEFEF}    &  \cellcolor[HTML]{EFEFEF}                                         & \multicolumn{3}{c|}{\cellcolor[HTML]{EFEFEF}HICO-Det} & \cellcolor[HTML]{EFEFEF}                         \\
\rowcolor[HTML]{EFEFEF} 
\multirow{-2}{*}{\cellcolor[HTML]{EFEFEF}$\lambda_1$} & \multirow{-2}{*}{\cellcolor[HTML]{EFEFEF}$\lambda_2$} & Full            & Rare            & Non-Rare          & \multirow{-2}{*}{\cellcolor[HTML]{EFEFEF}V-COCO} \\ \hline
% \multirow{2}{*}{$\lambda_1$} & \multirow{2}{*}{$\lambda_2$} & \multicolumn{3}{c|}{HICO-Det} & \multirow{2}{*}{V-COCO} \\
                                           % &                                            & Full    & Rare    & Non-Rare  &                         \\ \hline
1                                          & 1                                          & \textbf{34.63}   & \textbf{30.24}   & \textbf{35.95}     & \textbf{63.7}                    \\
1                                          & 0.1                                        & 34.38   & 29.93   & 35.82     & 63.4                    \\
0.1                                        & 1                                          & 34.20   & 29.92   & 35.49     & 63.2                    \\
0.1                                        & 0.1                                        & 34.21   & 29.84   & 35.58     & 63.1                    \\ \hline \specialrule{1.5pt}{0pt}{0pt}
\end{tabular}}
\captionof{table}{Ablation study using different $\lambda_1$ and $\lambda_2$ on HICO-Det datasets.}
\label{tab:ab_labmda}
\end{minipage}

\vspace{0.2cm}

% 第三行的两个表格
\begin{minipage}[t]{0.42\textwidth}
\centering
\setlength{\tabcolsep}{1.5mm}{%
\begin{tabular}{ll|ccc|c}
\hline \specialrule{1.5pt}{0pt}{0pt}
\rowcolor[HTML]{EFEFEF} 
 \cellcolor[HTML]{EFEFEF}    &  \cellcolor[HTML]{EFEFEF}                                         & \multicolumn{3}{c|}{\cellcolor[HTML]{EFEFEF}HICO-Det} & \cellcolor[HTML]{EFEFEF}                         \\
\rowcolor[HTML]{EFEFEF} 
\multirow{-2}{*}{\cellcolor[HTML]{EFEFEF}$\gamma_1$} & \multirow{-2}{*}{\cellcolor[HTML]{EFEFEF}$\gamma_2$} & Full            & Rare            & Non-Rare          & \multirow{-2}{*}{\cellcolor[HTML]{EFEFEF}V-COCO} \\ \hline
% \multirow{2}{*}{$\gamma_1$} & \multirow{2}{*}{$\gamma_2$} & \multicolumn{3}{c|}{HICO-Det} & \multirow{2}{*}{V-COCO} \\
                                           % &                                            & Full    & Rare    & Non-Rare  &                         \\ \hline
1                                          & 1                                          & \textbf{34.63}   & \textbf{30.24}   & \textbf{35.95}     & \textbf{63.7}                    \\
1                                          & 0.1                                        & 34.35   & 29.87   & 35.74     & 63.5                    \\
0.1                                        & 1                                          & 33.91   & 29.42   & 35.32     & 63.1                    \\
0.1                                        & 0.1                                        & 33.85   & 29.34   & 35.27     & 62.8                    \\ \hline \specialrule{1.5pt}{0pt}{0pt}
\end{tabular}}
\captionof{table}{Ablation study using different $\gamma_1$ and $\gamma_2$ on HICO-Det and V-COCO.}
\label{tab:ab_gamma}
\end{minipage}
\hfill
\begin{minipage}[t]{0.56\textwidth}
\centering
\small
\setlength{\tabcolsep}{0.5mm}{%
\begin{tabular}{c|ccc|c}
\hline \specialrule{1.5pt}{0pt}{0pt}
\rowcolor[HTML]{EFEFEF} 
 \cellcolor[HTML]{EFEFEF}                                           & \multicolumn{3}{c|}{\cellcolor[HTML]{EFEFEF}HICO-Det} & \cellcolor[HTML]{EFEFEF}                         \\
\rowcolor[HTML]{EFEFEF} 
\multirow{-2}{*}{\cellcolor[HTML]{EFEFEF}Templates} & Full            & Rare            & Non-Rare          & \multirow{-2}{*}{\cellcolor[HTML]{EFEFEF}V-COCO} \\ \hline
% \multirow{2}{*}{Templates} & \multicolumn{3}{c|}{HICO-Det} & \multirow{2}{*}{V-COCO} \\
               % & Full    & Rare    & Non-Rare  &                         \\ \hline
person \{1\} \{2\} & 34.61& 30.14& 35.75& 63.7 \\
someone \{1\} a/an \{2\} outdoors or indoors & 34.57& 30.03& 35.83&63.6\\
a person is \{1\}ing a/an \{2\} & \textbf{34.63}& \textbf{30.24}& \textbf{35.95}& \textbf{63.7}\\
person interacting with a/an \{2\} by \{1\}ing & 34.37& 29.89& 35.68&63.4\\ \hline \specialrule{1.5pt}{0pt}{0pt}
\end{tabular}}
\captionof{table}{Performance of different text templates in ACTOR on HICO-Det and V-COCO.}
\label{tab:templates}
\end{minipage}
% \vspace{-0.3cm}
\end{figure*}

\subsubsection{Training Efficiency Analysis}
Table~\ref{tab:training_efficiency} demonstrates that QueryCraft significantly accelerates training convergence. All evaluated methods show consistent epoch reductions: TED-Net (-18.6\%), LOGICHOI (-13.9\%), KI2HOI (-13.3\%), and GEN-VLKT (-12.6\%).  This acceleration stems from improved initialization landscapes—while random initialization requires models to simultaneously learn query semantics and detection patterns, our semantic initialization provides meaningful starting points through PDQD's object-awareness and ACTOR's action-relevant information. This reduces the hypothesis space during training, leading to faster convergence and more stable optimization.

% The practical benefits include reduced computational costs (TED-Net saves ~18 epochs), faster experimentation cycles, and better optimization trajectories. The consistent reduction rates across diverse methods confirm that efficiency gains are a fundamental benefit of semantic query initialization rather than method-specific artifacts, making QueryCraft a practical enhancement for real-world HOI detection deployment.

\subsection{Ablation Studies}

We conduct comprehensive ablation studies on HICO-DET and V-COCO datasets. All ablation experiments are performed using GEN-VLKT as the baseline method with ResNet-50 backbone.

\subsubsection{Component Analysis}
Table~\ref{tab:ab_commpoent} presents ablation results for ACTOR and PDQD modules. From the baseline of 33.51 mAP (GEN-VLKT without either module), enabling PDQD alone yields +0.32 mAP improvement, while ACTOR alone provides a larger +0.73 mAP gain, with particularly strong improvements on Rare interactions (+1.39 mAP). This suggests action semantics represent a primary bottleneck in HOI detection.
Crucially, combining both modules achieves 34.63 mAP (+1.12 total), exceeding the sum of individual gains (1.05), indicating positive synergy. 

\subsubsection{Impact of Weighting Parameters}
Tables~\ref{tab:ab_labmda} and \ref{tab:ab_gamma} analyze the weighting parameters for PDQD ($\lambda_1$, $\lambda_2$) and ACTOR ($\gamma_1$, $\gamma_2$). Both modules achieve optimal performance with all parameters set to 1.0. For PDQD, reducing $\lambda_1$ (initialization) to 0.1 causes a -0.43 mAP drop, while reducing $\lambda_2$ results in -0.25 mAP, indicating initialization is more critical. ACTOR shows even stronger sensitivity: reducing $\gamma_1$ to 0.1 yields a substantial -0.72 mAP decrease, nearly double PDQD's impact. The Rare subset is particularly sensitive to ACTOR parameters (30.24→29.34 mAP when both are 0.1), confirming that action semantics are crucial for rare interactions. 
% Key findings: (1) Both initialization and output enhancement stages contribute significantly, validating our dual-integration design; (2) ACTOR parameters have larger impact than PDQD, suggesting action understanding is the primary challenge in HOI; (3) The consistent optimal value of 1.0 indicates well-calibrated semantic representations that require no scaling.

\subsubsection{Impact of text Templates in ACTOR}
Table~\ref{tab:templates} evaluates four template variations for generating action descriptions in ACTOR. Overall, the impact of using different language prompts on model performance is negligible. The progressive template ``a person is \{1\}ing a/an \{2\}") achieves optimal performance (34.63 mAP), while the minimal template (``person \{1\} \{2\}") performs nearly identically (-0.02 mAP), demonstrating robustness to simplicity. 
% Adding spatial context (``outdoors or indoors'') or explicit interaction phrases ("interacting with...by") actually decreases performance (-0.06 and -0.26 mAP respectively), suggesting that unnecessary complexity dilutes essential action-object relationships.  
The small performance variations (maximum 0.26 mAP difference) across substantially different templates confirm ACTOR's robustness to linguistic variations, attributed to the vision-language model's ability to map diverse expressions to similar semantic representations.

\section{Conclusion}

In this paper, we presented QueryCraft, a novel framework that addressed the fundamental limitation of random query initialization in DETR-based HOI detection methods. Through comprehensive semantic query initialization, we demonstrated that providing meaningful priors significantly improves both detection performance and training efficiency. QueryCraft introduces two modules: PDQD leverages knowledge distillation from pre-trained object detectors to generate object-aware queries, and ACTOR exploits vision-language alignment to create action-semantic queries through cross-modal attention. These modules work synergistically, with PDQD providing robust object understanding and ACTOR enabling compositional reasoning about interactions. Extensive experiments validate the effectiveness of QueryCraft across multiple dimensions. 

\bibliography{aaai2026}

% \input{AnonymousSubmission/LaTeX/ReproducibilityChecklist}


\end{document}